\documentclass[conference]{IEEEtran}
\IEEEoverridecommandlockouts
\usepackage{cite}
\usepackage{amsmath,amssymb,amsfonts}
\usepackage{algorithmic}
\usepackage{graphicx}
\usepackage{textcomp}
\usepackage{xcolor}
\usepackage{lipsum} % For dummy text
\usepackage{subfigure}
\usepackage{subcaption}
\usepackage{caption}
\usepackage{multirow}
\usepackage{booktabs}
\usepackage{array} % For centering columns

\def\BibTeX{{\rm B\kern-.05em{\sc i\kern-.025em b}\kern-.08em
    T\kern-.1667em\lower.7ex\hbox{E}\kern-.125emX}}
\begin{document}

\title{AstroMAE: Redshift Prediction Using a Masked Autoencoder with a Novel Fine-Tuning Architecture\\
% {\footnotesize \textsuperscript{*}Note: Sub-titles are not captured in Xplore and
% should not be used}
% \thanks{Identify applicable funding agency here. If none, delete this.}
}

\author{\IEEEauthorblockN{Amirreza Dolatpour Fathkouhi}
\IEEEauthorblockA{\textit{Department of Computer Science} \\
\textit{University of Virginia}\\
Charlottesville, USA \\
aww9gh@virginia.edu}
\and
\IEEEauthorblockN{Geoffrey Charles Fox}
\IEEEauthorblockA{\textit{Department of Computer Science} \\
\textit{University of Virginia}\\
Charlottesville, USA \\
vxj6mb@virginia.edu}
}

\maketitle

\begin{abstract}
Redshift prediction is a fundamental task in astronomy, essential for understanding the expansion of the universe and determining the distances of astronomical objects. Accurate redshift prediction plays a crucial role in advancing our knowledge of the cosmos. Machine learning (ML) methods, renowned for their precision and speed, offer promising solutions for this complex task. However, traditional ML algorithms heavily depend on labeled data and task-specific feature extraction. To overcome these limitations, we introduce AstroMAE, an innovative approach that pretrains a vision transformer encoder using a masked autoencoder method on Sloan Digital Sky Survey (SDSS) images. This technique enables the encoder to capture the global patterns within the data without relying on labels. To the best of our knowledge, AstroMAE represents the first application of a masked autoencoder to astronomical data. By ignoring labels during the pretraining phase, the encoder gathers a general understanding of the data. The pretrained encoder is subsequently fine-tuned within a specialized architecture tailored for redshift prediction. We evaluate our model against various vision transformer architectures and CNN-based models, demonstrating the superior performance of AstroMAE’s pretrained model and fine-tuning architecture.

\end{abstract}


\section{Introduction}

Redshift prediction is one of the most compelling areas of study in astronomy, offering insights into the universe's expansion and the distances of celestial objects such as quasars, stars, and galaxies \cite{int1}. Accurately capturing spectral features over extended periods is crucial for redshift prediction, but this is feasible for only 1\% of galaxies. Additionally, most telescopes can simultaneously capture spectra from only a limited number of objects \cite{int2}. Consequently, photometric methods have been proposed as alternatives, since spectroscopic methods are both expensive and time-consuming \cite{int3}. Thanks to advancements in telescopes, a vast number of images are captured and made available by various surveys, such as DESI \cite{int5} and Hyper Suprime-Cam \cite{int6}.

Photometric redshift prediction can be achieved through two main approaches: template-fitting methods and machine learning-based methods, particularly deep learning. Template-fitting methods, such as those researched by Salvato et al. \cite{int4}, aim to determine the probability density function of redshift \cite{int9}. Given this paper's focus on deep learning, most reviews explore deep learning-based methods in astronomy, specifically for redshift prediction.

% Deep learning methods due to large number of parameters can solve more sophisticated problems compared to traditional method. In addition, development of hardware, particularly GPUs, give us

Deep learning methods proposed for astronomy can mainly be categorized into supervised and self-supervised learning algorithms.

In supervised learning, the model's training process is dependent on the availability of labeled data, where the target values guide the learning of task-specific features. Dey et al. \cite{int2} proposed a method based on capsule cells. Pasquet et al. \cite{m7} used an Inception model to extract features from images, which were then concatenated with galactic reddening values. Rastegarnia et al. \cite{int10} used residual blocks to predict quasar redshifts. In \cite{int11}, a deep learning model based on a convolutional neural network (CNN) was designed to predict quasar redshifts. Sandeep et al. \cite{int12}, in addition to using pretrained models such as AlexNet, VGG16, and ResNet50, proposed a new CNN-based model to classify galaxies and predict their redshifts. Syarifudin et al. \cite{int13} used multi-band images and a DenseNet model to predict redshifts. Schuldt et al. \cite{int14} proposed Netz, a CNN deep learning model trained on five-filter images collected from the Hyper Suprime-Cam Subaru Strategic Program. In \cite{intViT}, a Vision Transformer (ViT) model, based on the transformer architecture, was trained for galaxy classification.

Supervised learning methods require labels for training, and the extracted features are specifically related to the defined task. Additionally, labeled data is not widely available, and the extracted features often fail to capture the general patterns of the data. To address this, self-supervised learning methods have been proposed to leverage abundant data without relying on labels. These methods define a pretext task \cite{int15} that the model solves, such as image inpainting \cite{int16}, allowing the model to learn general patterns and features of the data.

Self-supervised learning includes two main phases. The first phase is pretraining, where the model is trained on the pretext task using unlabeled data to identify general patterns. The second phase is fine-tuning, where the pretrained model is used to solve a specific task. Hayat et al. \cite{int17} used contrastive learning for pretraining the ResNet50 model, with the pretrained weights later employed for morphology classification and redshift prediction. Lanusse et al. \cite{int18} proposed AstroCLIP, a multimodal model pretrained using a contrastive learning approach on the DESI survey, which was then used for redshift and stellar mass prediction. Shen et al. \cite{int19} pretrained ResNet50 with a momentum contrastive learning method, later employing the pretrained model for Galaxy Zoo classification. Stein et al. \cite{int20} pretrained ResNet50 for similarity search between galaxies. Oliva et al. \cite{int21} pretrained a transformer-based architecture using a masking strategy on millions of R-band light curves.

Previous studies mainly used various contrastive learning methods \cite{int22} for pretraining. These methods are highly sensitive to the selected augmentation techniques \cite{int23, int24}. Incorrect augmentations can significantly degrade model performance. Furthermore, with a greater number of different views, more patches are processed by the encoder. In contrast, Masked AutoEncoder (MAE) \cite{m4}, a self-supervised learning method, uses only 25\% of patches, making it more efficient compared to contrastive learning methods. According to experiments conducted in \cite{m4}, MAE is not sensitive to the type of augmentation techniques used. MAE employs a vision transformer \cite{m1} as the encoder, which can extract the global dependency and general patterns of images but lacks the ability to capture locality information, such as edges, which CNNs can intuitively extract.

To address the aforementioned issues, this paper makes the following contributions:
\begin{itemize}
\item We pretrained a model on a portion of the SDSS survey images by ignoring the labels, using a masked autoencoder approach to gather the general and global features of the data. This method aims to achieve faster, more efficient pretraining that is less sensitive to augmentation. To the best of our knowledge, this is the first paper to explore the usage of masked autoencoders for astronomy images.
\item To mitigate the lack of locality in vision transformer models, we introduce a novel fine-tuning method specifically designed for redshift prediction.
\item We design various architectures based on vision transformers and CNNs to address these issues and demonstrate the superiority of our proposed AstroMAE through different experiments.
\end{itemize}

The remainder of this paper is structured as follows: Section \ref{sec:II} delves into the AstroMAE architecture and outlines the proposed fine-tuning methods. Section \ref{sec:III} presents our experiments, detailing both the pretraining and fine-tuning processes, along with a comprehensive analysis of our results. Ultimately, we conclude with a summary of our findings and the implications of our experiments.

\begin{figure}[h]
    \centering
    \begin{minipage}[b]{0.3\columnwidth}
        \centering
        \includegraphics[width=\linewidth]{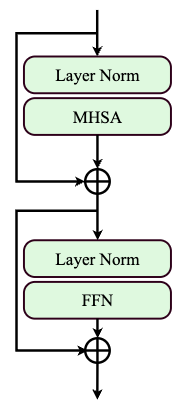}
        \caption*{(a)}
    \end{minipage}
    % \hfill
    \begin{minipage}[b]{0.46\columnwidth}
        \centering
        \includegraphics[width=\linewidth]{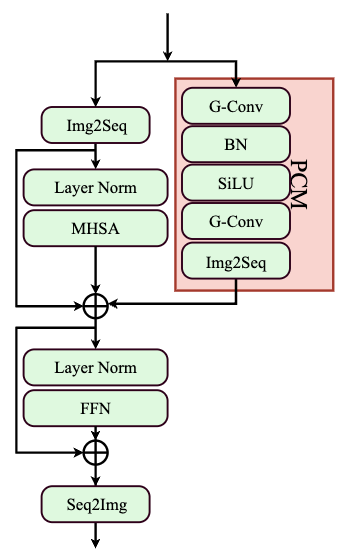}
        \caption*{(b)}
    \end{minipage}
    \caption{These diagrams illustrate (a) a plain-transformer and (b) a pcm-transformer. The abbreviations used are as follows: Layer Norm (Layer Normalization), MHSA (Multi-Head Self-Attention), FFN (Feed Forward Network), and BN (Batch Normalization). Img2Seq and Seq2Img represent the processes of converting between 1D and 2D features. G-Conv denotes a group of convolutional layers, and SiLU layer is explained in \cite{silu}.}
    \label{fig:plain_pcm_transformer}
\end{figure}

\section{Proposed Method}
\label{sec:II}
This section begins with an introduction to the architectures and concepts utilized in this paper. Then, the proposed model architectures are presented.

\subsection{Vision Transformer}
The Vision Transformer (ViT), proposed by Dosovitskiy et al. \cite{m1}, aims to train images using a plain transformer layer. It begins by segmenting an image $x \in R^{H \times W \times C}$ into uniform patches $x_t \in R^{\frac{(H\times W)}{p^2} \times D}$, where H, W, and C represent the height, width, and channel of the image x, and p represents the size of each patch, respectively. These patches are then transformed into embedding vectors of size $D = p^{2}C$ via a linear projector. A learnable class token is also concatenated with these patch embedding vectors. Subsequently, positional embedding vectors are added pairwise to the patch embeddings to inform the transformer layers about their positions in the image.

The plain transformer layer includes multi-head self-attention (MHSA) and a feed-forward network (FFN).

\textbf{MHSA}: Each token $x_t$ is projected to query (Q), key (K),and Value (V) vectors for \textit{h} times using $W_Q, W_K, W_V \in R^{D \times D}$.
\begin{gather}
    Q = W_Q \times x_t \\
    K = W_K \times x_t \\
    V = W_V \times x_t 
\end{gather}
For each of these projections, self-attention is conducted as demonstrated below:
\begin{gather}
    Attention(Q, K, V) = Softmax(\frac{QK^{T}}{\sqrt{D}})V
\end{gather}
The output of the MHSA is the concatenation of all calculated \textit{Attention} outputs along the channel dimension.

\textbf{FFN:} This module contains two linear layers with a GeLU activation function \cite{gelu} added between them.

As shown in Fig. \ref{fig:plain_pcm_transformer}, in each plain transformer layer, the input is normalized before being fed to each module, and a residual connection is added afterward.

\subsection{Transformer Layer with Parallel Convolution Module}
One drawback of the Vision Transformer (ViT) is that it cannot capture the local information of the image, which could be intuitively obtained by a Convolutional Neural Network (CNN). In ViT, images are treated as sequences of 1D tokens, and the 2D structure of images is not retained during training. In contrast, CNN applies kernel operators directly to images, effectively capturing the correlation between pixels. The Multi-Head Self Attention (MHSA) module of the plain transformer can model global dependencies by calculating the attention, but only the Feed-Forward Network (FFN) module is dedicated to capturing local information \cite{m1}.

To address this issue, a modified version of ViTAE \cite{m2} leverages both CNN and plain transformer layers \cite{m3}. In this method, a parallel convolution-based module (PCM) is utilized and integrated with the output of the attention module. Consequently, the inductive bias obtained by the CNN is combined with the global dependency captured by the MHSA. Fig. \ref{fig:plain_pcm_transformer} illustrates the modified transformer layer proposed in \cite{m3}. In this paper, we refer to the modified transformer as the pcm-transformer.

\begin{figure}[h]
  \centering
  \includegraphics[width=\columnwidth]{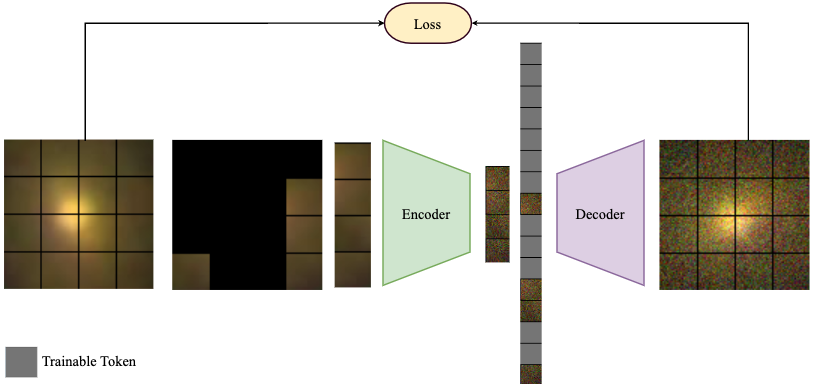}
  \caption{It illustrates the architecture of pretraining the AstroMAE using a masked autoencoder algorithm. Loss involves comparing the generated patches corresponding to masked areas with their original counterparts.}
  \label{fig:masked_autoencoder}
\end{figure}

\subsection{Masked AutoEncoder}

The Masked Autoencoder (MAE), a pre-training technique developed by Facebook \cite{m4}, is an asymmetrical autoencoder designed to extract patterns from data by reconstructing original images from portions of those images. The MAE comprises an encoder and a decoder. The encoder is a Vision Transformer (ViT) that features transformer-based layers, while the decoder can be constructed using either transformer or linear layers \cite{m5}.

The process begins with segmenting images into uniform patches, which are subsequently transformed into embedding vectors via a convolution-based patch embedder. Positional embeddings—computed using sine-cosine functions—are integrated into the patch embeddings to inform the encoder about the position of each patch. A predetermined mask ratio dictates the random removal of certain patches, with the remaining patches then fed into the encoder. To make the problem more challenging for the MAE and avoid using extrapolation to predict masked patches from neighboring pixels, a high masking ratio is typically used (75\%). In place of the masked patches, learnable tokens are generated and combined with the output of the encoder. These tokens are further enhanced with positional embeddings before being forwarded to the decoder. Ultimately, the decoder attempts to reconstruct the original image, and the model's effectiveness is assessed based on how accurately the reconstructed patches match their corresponding original segments. By masking patches and feeding only a small fraction of them, the pre-training process becomes significantly faster and more efficient. The MAE architecture is illustrated in Fig. \ref{fig:masked_autoencoder}.

\subsection{Inception module}
\begin{figure}[h]
  \centering
  \includegraphics[width=\columnwidth]{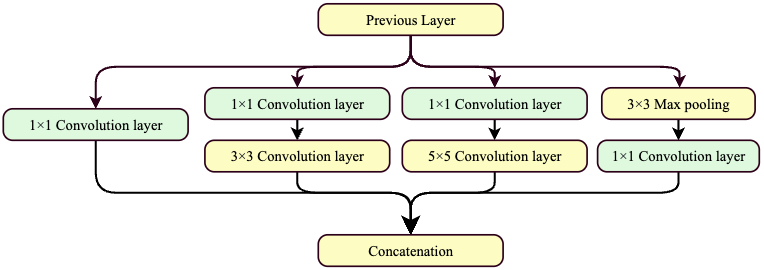}
  \caption{Inception module}
  \label{fig:inception_module}
\end{figure}

The Inception module \cite{m6} is constructed using CNN. In this architecture, four parallel branches of convolutions are integrated to increase both the depth and the width of the model. To make the process more efficient and decrease computations, a convolution layer with a kernel size of $1 \times 1$ is added before convolutions with sizes $3 \times 3$ and $ 5 \times 5$. The Inception module architecture is shown in Fig. \ref{fig:inception_module}.

\subsection{AstroMAE (Proposed Method)}
AstroMAE, involves using a masked autoencoder for pretraining and utilizing the pretrained encoder in a novel fine-tuning architecture designed for redshift prediction.

\textbf{Pretraining:}
AstroMAE is pretrained using a masked autoencoder. We employed both plain-transformer and pcm-transformer layers for constructing the AstroMAE. We pretrained two versions of AstroMAE: one based on the plain-transformer layer, called plain-AstroMAE, and another using pcm-transformer layers, called pcm-AstroMAE.

\textbf{Fine-tuning:}
The proposed fine-tuning model, depicted in Fig. \ref{fig:second_three_architectures}, contains three separate modules explained below:

\paragraph{Pretrained Encoder}
The decoder part of the AstroMAE is discarded, and only the encoder is used for fine-tuning. Two linear layers with a ReLU activation function between them serve as the head of the encoder. Additionally, fine-tuning is done partially, meaning that all weights of the pretrained encoder, except for those in the head, are frozen.

\paragraph{Inception Model}
\label{sec:inception}
This branch of the architecture contains five inception blocks, as explained earlier. The first four blocks include all four parallel branch convolution layers. However, in the last inception block, the branch containing the convolution layer with a kernel size of $5 \times 5$ is omitted. This is because the input to this branch is too small to apply a convolution with a $5 \times 5$ kernel. 
\paragraph{Magnitude Block}
This block includes a multi-layer perceptron comprising five linear layers with ReLU activation functions between them.

As depicted in Fig. \ref{fig:second_three_architectures}, the proposed fine-tuning architecture consists of the concatenation of the Inception model, the magnitude block, and the frozen pretrained encoder, which are then fed into two linear layers with a ReLU activation function between them.

\begin{figure}[h]
  \centering
  \includegraphics[width=\columnwidth]{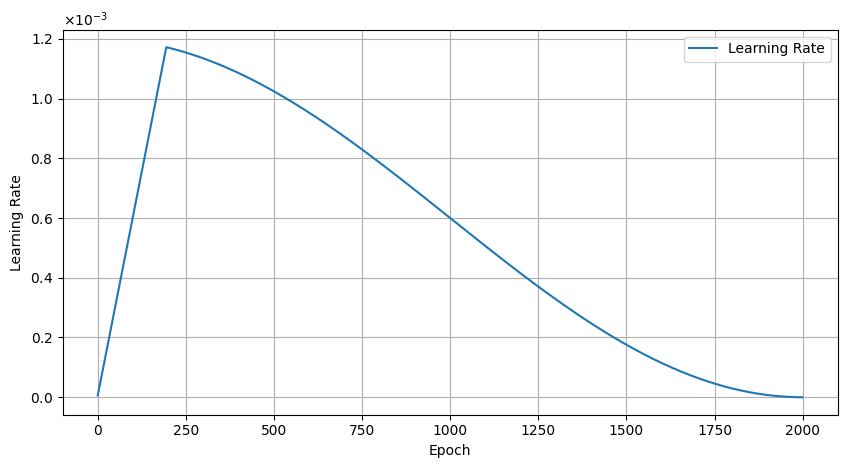}
  \caption{Learning rate during pretraining.}
  \label{fig:pretrained_lr}
\end{figure}

\begin{figure}[h]
  \centering
  \includegraphics[width=\columnwidth]{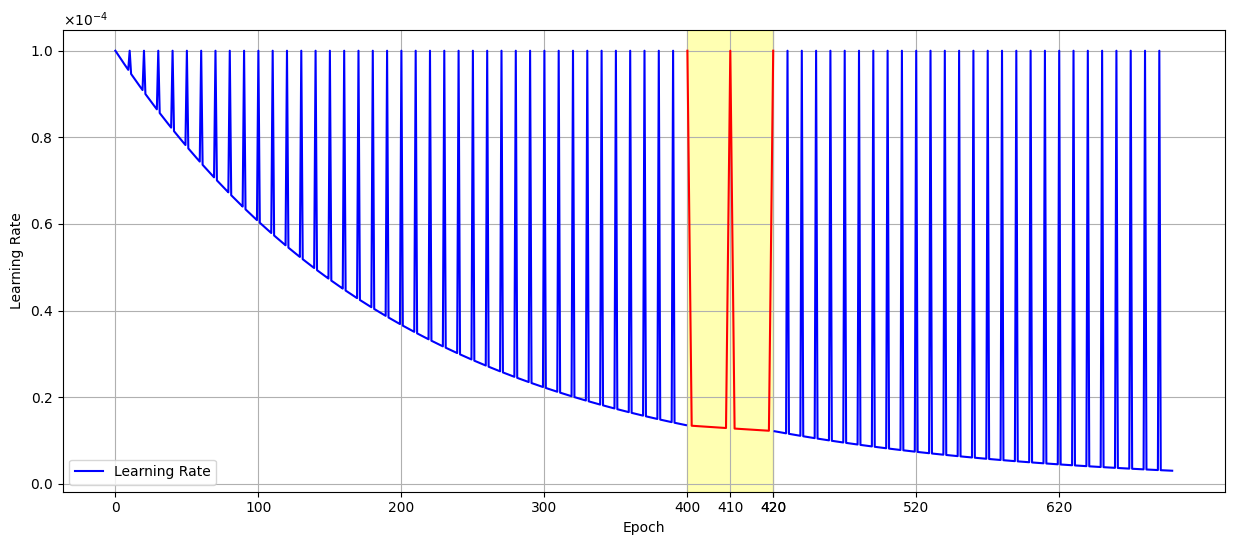}
  \caption{It shows the learning rate during fine-tuning, with the yellow section highlighting the changes over two cycles.}
  \label{fig:finetuned_lr}
\end{figure}

\begin{table}[h]
    \centering
    \renewcommand{\arraystretch}{1.3} % Adjust the row height as needed
    \caption{Fine-tuning Hyperparameters}
    \begin{tabular}{ll}
        \toprule
        \textbf{Hyperparameter} & \textbf{Value} \\
        \midrule
        $lr_{initial}$ & 1e-4 \\
        batch size & 1,024 \\
        seed & 42 \\
        total epochs & 700 \\
        optimizer & AdamW \cite{adamw} \\

        weight decay & 0.005 \\
        betas & (0.9, 0.999) \\
        \bottomrule
    \end{tabular}
    \label{tab:finetuning_hyperparameters}
\end{table}

\begin{table}[h]
    \centering
    \renewcommand{\arraystretch}{1.3} % Adjust the row height as needed
    \caption{Pretraining Hyperparameters}

    \begin{tabular}{ll}
        \toprule
        \textbf{Hyperparameter} & \textbf{Value} \\
        \midrule
        $lr_{peak}$ & 1.17e-3 \\
        batch size & 2,048 \\
        mask ratio & 0.75 \\
        seed & 42 \\
        total epochs & 2,000 \\
        optimizer & AdamW \\
        $epoch_{warm-up}$ & 196 \\
        weight decay & 0.05 \\
        betas & (0.9, 0.95) \\
        \bottomrule
    \end{tabular}
    \label{tab:pretraining_hyperparameters_}
\end{table}

\section{Experimental Results}
\label{sec:III}
We conducted two experimental setups, utilizing 80\% and 100\% of the image dataset for pretraining in the first and second experiments, respectively. In the second experiment, we compared the best-performing AstroMAE architecture from the first experiment, pcm-AstroMAE, with the baseline method proposed by Henghes et al. \cite{m8}. This comparison was conducted to assess the generality and robustness of the proposed methods when applied to a larger dataset, ensuring that the performance gains observed with 80\% of the data scale effectively with the full dataset.\\\\
\textbf{Implementation: }PyTorch, one of the well-known implementation frameworks, was utilized for this study. Special thanks to Rivanna High-Performance Computing for providing the necessary computational resources. Pretraining was conducted using four A100 GPUs, typically taking two to three days to complete. Fine-tuning experiments were performed based on GPU availability, utilizing either one or four A100 GPUs. The initial fine-tuning took approximately 1 hour using one GPU, while fine-tuning the full set of labeled data typically took around 10 hours when using four GPUs.\\\\
\textbf{Dataset:}
The dataset provided by Pasquet et al. \cite{m7} contains 659,857 images with 64 corresponding physical properties. Physical properties, such as spectroscopic redshift z are collected from the 12th version of the Sloan Digital Sky Survey (SDSS DR12) \cite{d1, d2, d3}. Images corresponding to these physical properties are retrieved from the DR8 SDSS survey. The images contain five bands, including u, g, r, i, and z frames, and the size of each image is $64 \times 64 \times5$. All raw images are preprocessed by background subtraction and the same zero-point photometric calibration. More information related to the dataset and preprocessing steps is explained in \cite{m7}.
\subsection{First Experiment:}

\textbf{Pretraining Data:}
As mentioned before, only images are
utilized for pretraining. Therefore, we ignored around 80\%
of the labels from the complete dataset to gather the global dependencies of the data and capture non-specific patterns. Consequently, 527,886 images are dedicated to the training data. To monitor the behavior of the model during pretraining, approximately 10\% of the whole dataset is set aside for validation. Similar to the training data, the labels of the validation set are ignored.\\
\textbf{Fine-tuning:}
The fine-tuning data consists of images along with their corresponding magnitude values, including u, g, r, i, and z, in addition to spectroscopic redshift z as the target, representing 10\% of the entire dataset. Moreover, the u, g, r, i, and z magnitude values are obtained using the astroquery library \cite{d4}. The data distribution for training, validation, and testing comprises 70\%, 10\%, and 20\% of the fine-tuning data, respectively. The fine-tuning data is also used for models trained from scratch, as shown in \ref{tab:performance_metrics}.

As mentioned in \cite{m8}, there is no significant difference between the results for images of size $32 \times 32 \times 5$ and those of size $64 \times 64 \times 5$. Based on this, images are cropped from the center to a size of $32 \times 32 \times 5$ during both pretraining and fine-tuning. To increase the difficulty and prevent overfitting, random rotation at 45 degrees, along with horizontal and vertical flipping methods, are applied to the training images during both the pretraining and fine-tuning phases. Additionally, Gaussian noise with a standard deviation of 0.05 is used during fine-tuning.

\begin{table}[h]
    \centering
    \renewcommand{\arraystretch}{1.3} % Adjust the row height as needed
    \caption{AstroMAE Pretraining Architectures}
    \begin{tabular}{llc}
        \toprule
        \textbf{Parameter} & \textbf{Component} & \textbf{Value} \\
        \midrule
        \multirow{2}{*}{patch size} & Encoder & 8 \\
         & Decoder & 8\\
        \midrule
        \multirow{2}{*}{embedding size} & Encoder & 192 \\
         & Decoder & 192\\
        \midrule
        \multirow{2}{*}{depth} & Encoder & 12 \\
         & Decoder & 4\\
        \midrule
        \multirow{2}{*}{number of heads} & Encoder & 3 \\
         & Decoder & 3\\
        \bottomrule
    \end{tabular}
    \label{tab:pretraining_architecture}
\end{table}

\subsubsection{Training Configurations} 
\textbf{\\Learning Rate Scheduler:}
In previous deep learning training, models were trained using a constant learning rate. This method can cause the model to underperform and is not effective for optimizing deep models. On this account, two schedulers are employed for both pretraining and fine-tuning. Both pretraining and fine-tuning learning rates $lr$  are demonstrated in Fig. \ref{fig:pretrained_lr}, \ref{fig:finetuned_lr}. Additionally, optimization details are mentioned in Table \ref{tab:finetuning_hyperparameters}, \ref{tab:pretraining_hyperparameters_}.

\begin{figure*}[t]
    \centering
    \begin{minipage}[b]{0.45\columnwidth}
        \centering
        \includegraphics[width=\linewidth]{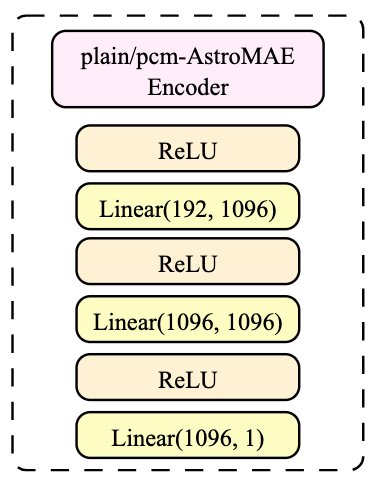}
        \caption*{(a)}
    \end{minipage}
    \hfill
    \begin{minipage}[b]{0.45\columnwidth}
        \centering
        \includegraphics[width=\linewidth]{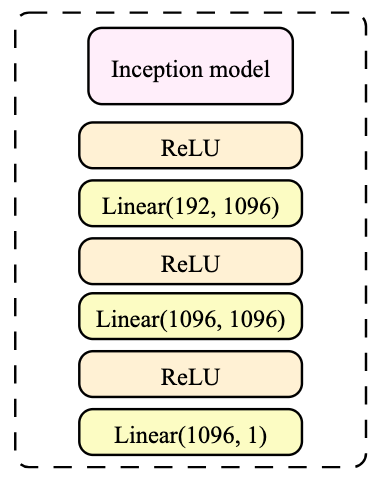}
        \caption*{(b)}
    \end{minipage}
    \hfill
    \begin{minipage}[b]{0.3\columnwidth}
        \centering
        \includegraphics[width=\linewidth]{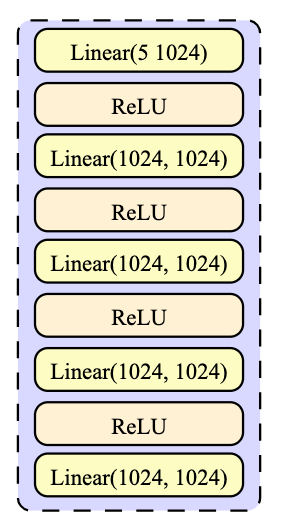}
        \caption*{(c)}
    \end{minipage}
    \hfill
    \begin{minipage}[b]{0.46\columnwidth}
        \centering
        \includegraphics[width=\linewidth]{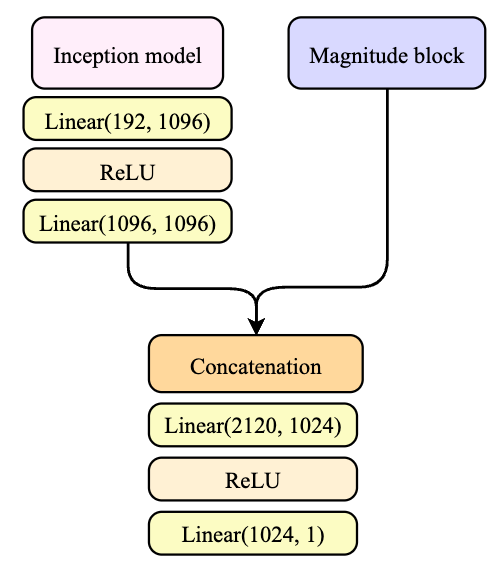}
        \caption*{(d)}
    \end{minipage}
    \caption{(a) plain-ViT, pcm-ViT, from-scratch plain-ViT, and from-scratch pcm-ViT, (b) Inception-only redshift prediction, (c) Magnitude Block, and (d) Henghes et al. \cite{m8} model.}
    \label{fig:first_architecture}
\end{figure*}

\begin{figure*}[t]
    \centering
    \begin{minipage}[b]{0.6\columnwidth}
        \centering
        \includegraphics[width=\linewidth]{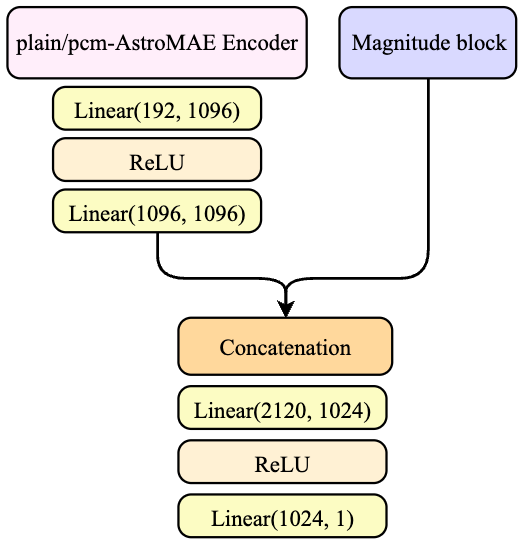}
        \caption*{(a)}
    \end{minipage}
    % \hfill
    \begin{minipage}[b]{0.6\columnwidth}
        \centering
        \includegraphics[width=\linewidth]{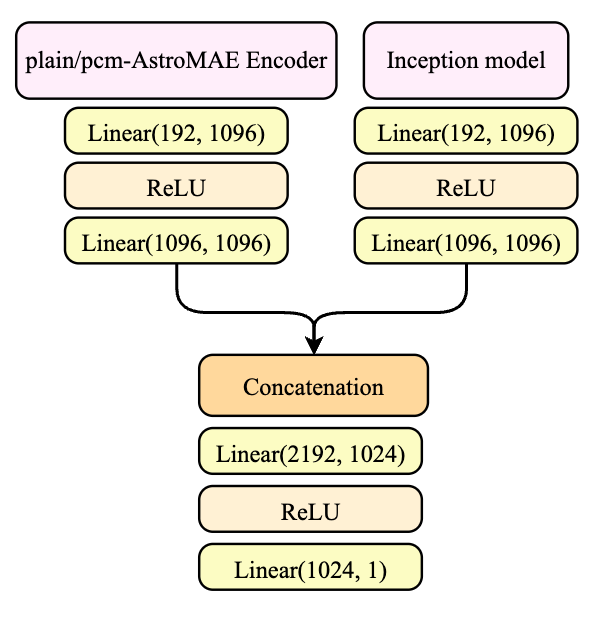}
        \caption*{(b)}
    \end{minipage}
    \begin{minipage}[b]{0.8\columnwidth}
        \centering
        \includegraphics[width=\linewidth]{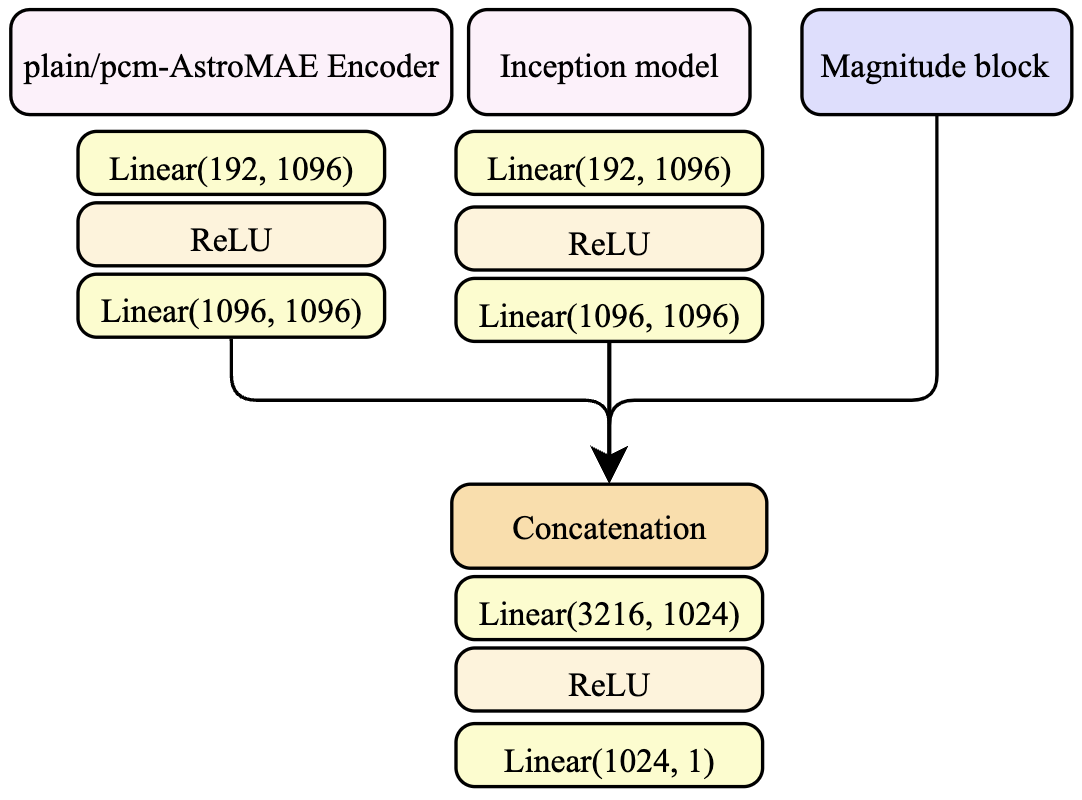}
        \caption*{(c)}
    \end{minipage}
    \caption{(a) plain-ViT-magnitude, pcm-ViT-magnitude, from-scratch plain-ViT-magnitude, and from-scratch pcm-ViT-magnitude, (b) plain-ViT-inception and pcm-ViT-inception, (c) Proposed AstroMAE Fine-tuning Architecture.}
    \label{fig:second_three_architectures}
\end{figure*}

\paragraph{Pretraining}
Similar to \cite{TC1}, a Cosine Annealing with Warm-Up scheduler is used for pretraining. This scheduler consists of two phases. In the first phase, the learning rate increases linearly from a low value to a high learning rate, $lr_{peak}$, over a specific number of epochs, called $epoch_{warm-up}$. In the second phase, the scheduler decreases the learning rate using cosine decay. Linearly increasing the learning rate avoids unstable training and improves the global search of the optimizer, while the cosine decay decreases the learning rate more smoothly, providing a good balance between global and local search.

\paragraph{Fine-tuning}
A cyclic scheduler is employed that restarts the $lr$ every 10 epochs, then decreases exponentially using $(0.995)^{epoch}$. This strategy helps the model escape local minima and achieve good convergence in the final epochs.

\textbf{AstroMAE Hyperparameters:}
For pretraining, compared to other papers that utilized MAE, our data is limited and not as large. Based on the analysis conducted in \cite{TC2}, larger models require training on larger datasets for a higher number of epochs. Therefore, we built small models by setting configurations mentioned in Table \ref{tab:pretraining_architecture} for both plain-AstroMAE and pcm-AstroMAE.

For fine-tuning, it is important to note that the encoder cannot be used directly as some shuffling is applied to patches during MAE pretraining. Consequently, the weights of the encoder should be extracted and used to initialize a new ViT model for fine-tuning. During fine-tuning, all layers except the last two layers of the ViT encoder (the layer normalization \cite{TC3} and projection layer) are frozen.

\subsection{Compare with other redshift predition methods}
To demostrate the superiority  of AstroMAE and the proposed fine-tuning architecture, we compared them with other redshift prediction model, which is based on the vision transformers or CNNs. All architectures are illustrated in Fig. \ref{fig:first_architecture}, \ref{fig:second_three_architectures}.

% \textbf{From-scratch training}
\textbf{plain-ViT and pcm-ViT:} The pretrained encoders of plain-AstroMAE and pcm-AstroMAE are employed for fine-tuning. A lightweight trainable head module is added at the end of the encoders for redshift prediction. In these architectures, the pretrained encoders are frozen during fine-tuning. 

\textbf{from-scratch plain-ViT and from-scratch pcm-ViT:} The architectures are the same as plain-ViT and pcm-ViT, but the encoders are initialized randomly and are trainable during training. Additionally, no pretraining is conducted on these architectures.

\textbf{Inception-only redshift prediction:} Similar to the Inception model discussed in the fine-tuning model architecture, it includes five inception modules, with the last one lacking the $5 \times 5$ convolution layer. Finally, three linear layers with ReLU activations between them are added. All weights of this model are trainable.

\textbf{Henghes et al. \cite{m8}:} In this paper, the Inception model is concatenated with the magnitude blocks. The output is then fed to two linear layers, with a ReLU activation function inserted between them.

\textbf{plain-ViT-inception and pcm-ViT-inception:} Plain-ViT and pcm-ViT are concatenated with the Inception model. Two linear layers with one ReLU function are then used for redshift prediction. Except for plain-ViT and pcm-ViT, all weights of the models are trainable.

\textbf{plain-ViT-magnitude and pcm-ViT-magnitude:} The magnitude block output is concatenated with the plain-ViT or pcm-ViT output before being fed to the linear layers.

\textbf{from-scratch plain-ViT-magnitude and from-scratch pcm-ViT-magnitude:} In this architecture, the magnitude block is only concatenated with the output of from-scratch plain-ViT and from-scratch pcm-ViT.

\begin{table*}[t]
    \centering
    \renewcommand{\arraystretch}{1.3} % Adjust the row height as needed
    \caption{Redshift Prediction Using Various Architectures Based on Transformer Layers and CNNs}
    \begin{tabular*}{\textwidth}{@{\extracolsep{\fill}} >{\centering\arraybackslash}p{1cm} >{\centering\arraybackslash}p{6cm} c c c c c}
        \toprule
        \multicolumn{2}{c}{\textbf{Architectures}} & \multicolumn{5}{c}{\textbf{Metrics}} \\
        \cmidrule(r){1-2} \cmidrule(l){3-7}
        \textbf{Type} & \textbf{Name} & \textbf{MSE} & \textbf{MAE} & \textbf{Bias} & \textbf{Precision} & \textbf{R\textsuperscript{2}} \\
        \midrule
        \multirow{5}{*}{\parbox[c]{3cm}{\centering Supervised training \\ (from scratch)}} & from-scratch plain-ViT-magnitude & 0.00077 & 0.01871 & 0.00153 & 0.01736 & 0.93580\\
        & from-scratch pcm-ViT-magnitude & 0.00057 & 0.01604 & -0.00035 & 0.01458 & 0.95204\\
        & Henghes et al. \cite{m8} & 0.00058 & 0.01568 & 0.00108 & 0.01443 & 0.95176 \\
        & from-scratch plain-ViT & 0.00097 & 0.02123 & 0.00049 & 0.01957 & 0.91871 \\
        & from-scratch pcm-ViT & 0.00063 & 0.01686 & -0.00122 & 0.01554 & 0.94764 \\
        & Inception-only redshift prediction & 0.00064 & 0.01705 & 0.00132 & 0.01593 & 0.94625 \\
        \midrule
        \multirow{6}{*}{\parbox[c]{3cm}{\centering Fine-tuning}} & plain-ViT-magnitude & 0.00068 & 0.01740 & \textbf{-0.00007} & 0.01596 & 0.94334 \\
        & pcm-ViT-magnitude & 0.00060 & 0.01655 & -0.00095 & 0.01522 & 0.94939 \\
        & Proposed plain-AstroMAE & 0.00056 & 0.01558 & 0.00097 & 0.01429 & 0.95336 \\
        & Proposed pcm-AstroMAE & \textbf{0.00053} & \textbf{0.01520} & -0.00037 & \textbf{0.01391} & \textbf{0.95601} \\

        & plain-ViT & 0.00086 & 0.01970 & -0.00060 & 0.01775 & 0.92790 \\
        & pcm-ViT & 0.00084 & 0.01945 & -0.00114 & 0.01737 & 0.92950 \\
        & plain-ViT-inception & 0.00059 & 0.01622 & -0.00009 & 0.01496 & 0.95029 \\
        & pcm-ViT-inception & 0.00059 & 0.01601 & 0.00042 & 0.01458 & 0.95095 \\
        \bottomrule
    \end{tabular*}
    \label{tab:performance_metrics}
\end{table*}

\subsection{Metrics}
Five metrics are considered for evaluating and comparing our proposed methods with others. These metrics are explained below:

\textbf{Mean Square Error (MSE):} The average of the squared differences between the spectroscopic and predicted redshift values is calculated.

\begin{equation}
\text{MSE} = \frac{1}{n} \sum_{i=1}^{n} (z^s_i - \hat{z}^s_i)^2
\end{equation}

\textbf{Mean Absolute Error (MAE):} The absolute differences between the predicted and ground-truth spectroscopic redshifts are averaged. 

\begin{equation}
    \text{MAE} = \frac{1}{n} \sum_{i=1}^{n} \left| z^s_i - \hat{z}^s_i \right|
\end{equation}

\textbf{Bias:} It measures the average of the residuals, as defined in \cite{mt1}.

\begin{equation}
    \text{Bias} = \langle \frac{\hat{z}^s - z^s}{1 + z^s} \rangle
\end{equation}

\textbf{Precision:} As mentioned in \cite{mt2}, it measures the expected scatter.

\begin{equation}
    \text{Precision} = 1.48 \times \text{median}(|\frac{\hat{z}^s - z^s}{1 + z^s}|).
\end{equation}

\textbf{$\mathbf{R^2}$ score:} It evaluates how well a regression model predicts. The $R^2$ score lies between 0 and 1, and the closer the score is to 1, the better the model predicts.

\begin{equation}
    R^2(z^s, \hat{z}^s) = 1 - \frac{\sum_{i=1}^{n} (z^s_i - \hat{z}^s_i)^2}{\sum_{i=1}^{n} (z^s_i - \bar{z}^s)^2}.
\end{equation}

In the above formulas, $z^s$, $\hat{z}^s$, and $\bar{z}^s$ represent the ground-truth spectroscopic redshift, predicted redshift, and average value of the spectroscopic redshift, respectively. Moreover, $n$ is the number of data samples. It is worth noting that methods with lower MSE, MAE, Bias, and Precision, and higher $R^2$ indicate better results.

\subsection{Result Analysis}
\begin{figure*}[t]
    \centering
    \begin{minipage}[b]{0.6\columnwidth}
        \centering
        \includegraphics[width=\linewidth]{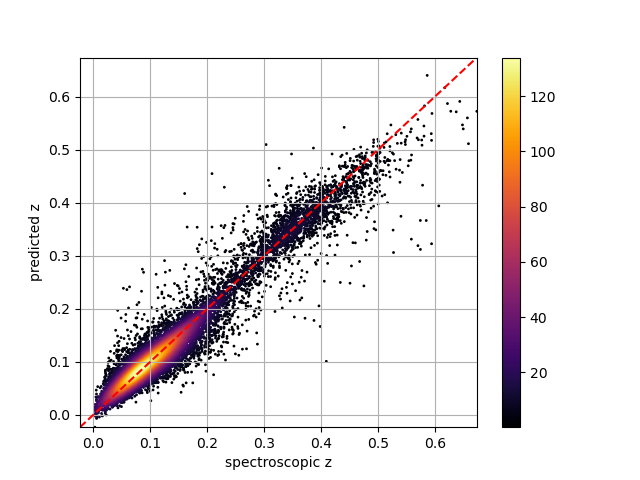}
        \caption*{(a)}
    \end{minipage}
    \begin{minipage}[b]{0.6\columnwidth}
        \centering
        \includegraphics[width=\linewidth]{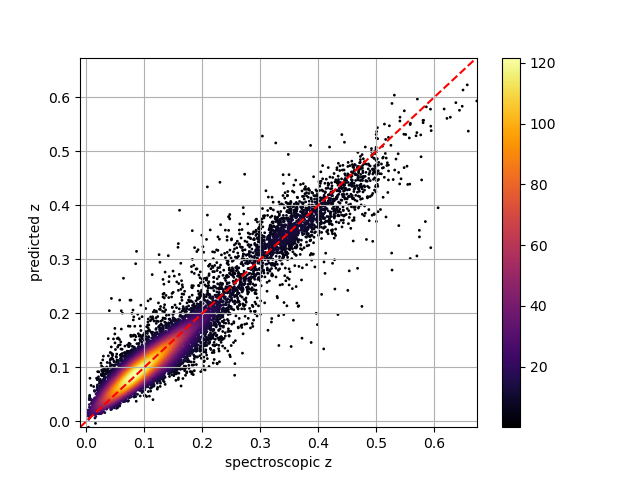}
        \caption*{(b)}
    \end{minipage}
    \begin{minipage}[b]{0.6\columnwidth}
        \centering
        \includegraphics[width=\linewidth]{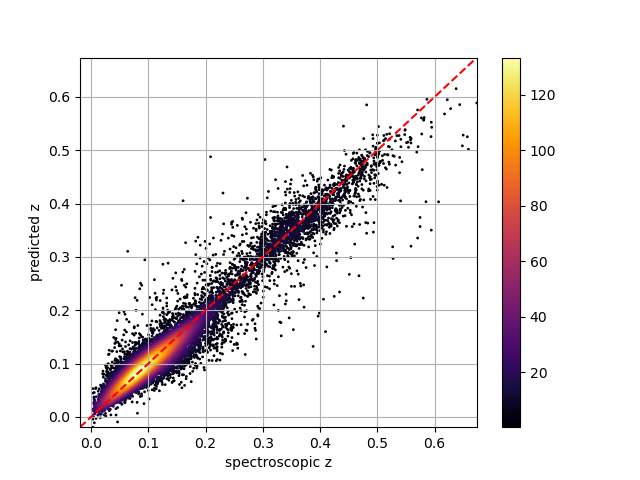}
        \caption*{(c)}
    \end{minipage}
    \begin{minipage}[b]{0.6\columnwidth}
        \centering
        \includegraphics[width=\linewidth]{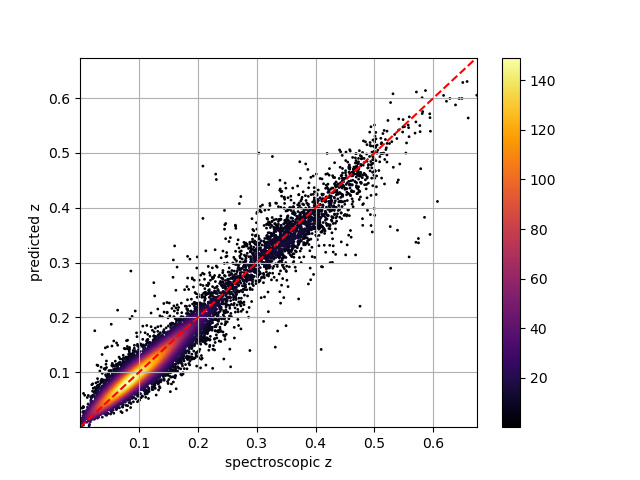}
        \caption*{(d)}
    \end{minipage}
    \begin{minipage}[b]{0.6\columnwidth}
        \centering
        \includegraphics[width=\linewidth]{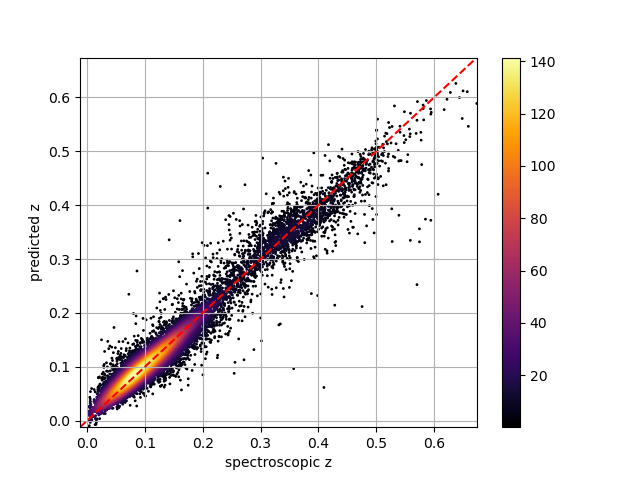}
        \caption*{(e)}
    \end{minipage}
    \begin{minipage}[b]{0.6\columnwidth}
        \centering
        \includegraphics[width=\linewidth]{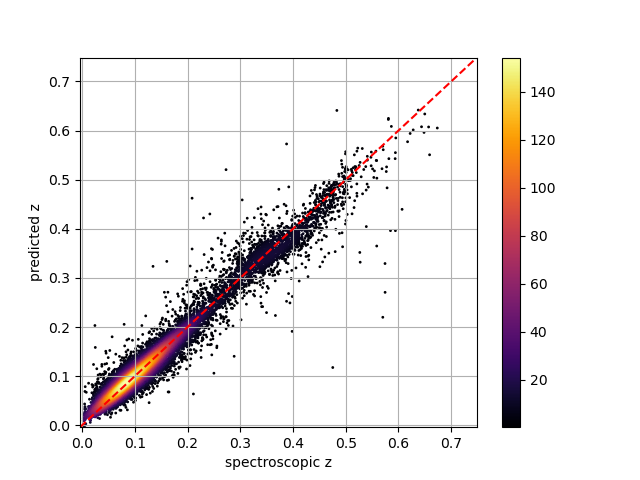}
        \caption*{(f)}
    \end{minipage}
    
    \caption{(a) plain-ViT, (b) from-scratch plain-ViT, (c) pcm-ViT (d) from-scratch pcm-ViT, (e) Inception-only redshift prediction, and (f) Henghes et al. \cite{m8} model.}
    \label{fig:first_six_density_scattler}
\end{figure*}

\begin{figure*}[t]
    \centering
    \begin{minipage}[b]{0.6\columnwidth}
        \centering
        \includegraphics[width=\linewidth]{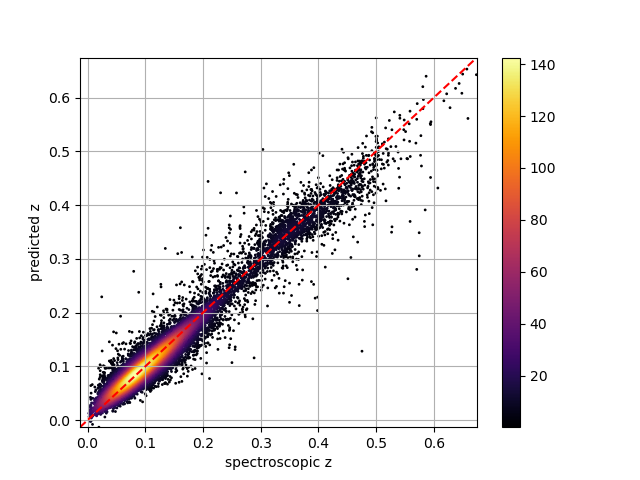}
        \caption*{(a)}
    \end{minipage}
    \begin{minipage}[b]{0.6\columnwidth}
        \centering
        \includegraphics[width=\linewidth]{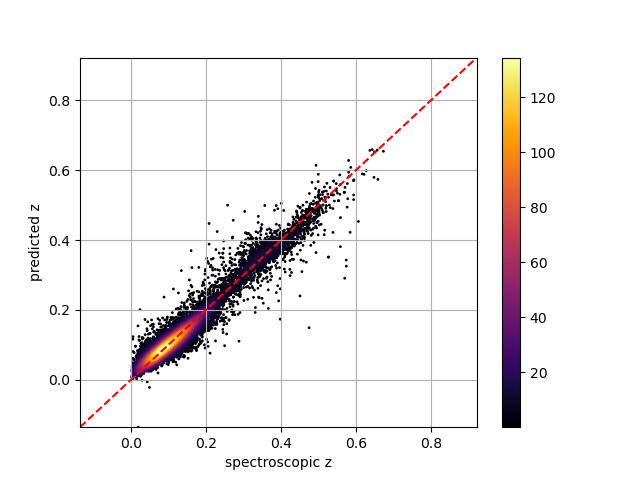}
        \caption*{(b)}
    \end{minipage}
    \begin{minipage}[b]{0.6\columnwidth}
        \centering
        \includegraphics[width=\linewidth]{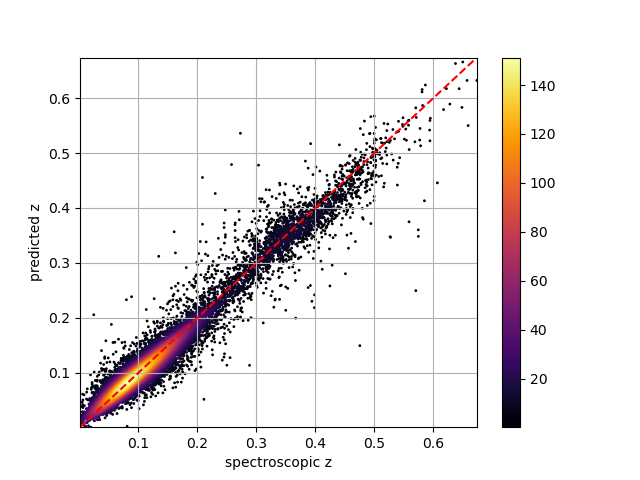}
        \caption*{(c)}
    \end{minipage}
    \begin{minipage}[b]{0.6\columnwidth}
        \centering
        \includegraphics[width=\linewidth]{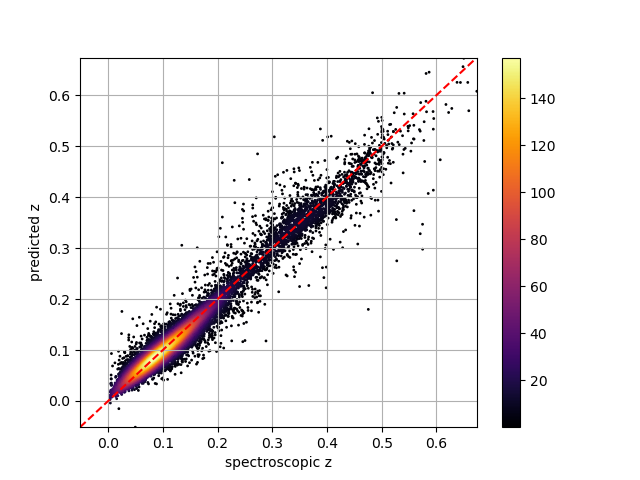}
        \caption*{(d)}
    \end{minipage}
    \begin{minipage}[b]{0.6\columnwidth}
        \centering
        \includegraphics[width=\linewidth]{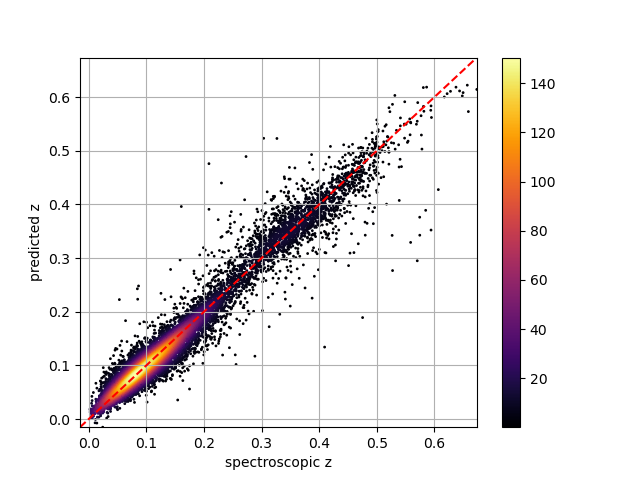}
        \caption*{(e)}
    \end{minipage}
    \begin{minipage}[b]{0.6\columnwidth}
        \centering
        \includegraphics[width=\linewidth]{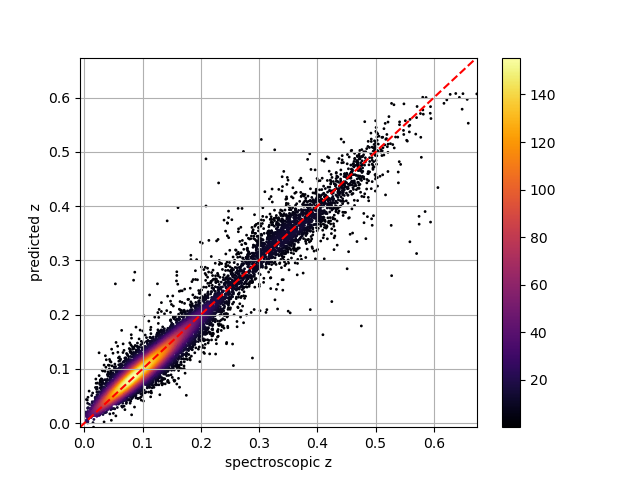}
        \caption*{(f)}
    \end{minipage}
    
    \caption{(a) plain-ViT-magnitude, (b) from-scratch plain-ViT-magnitude, (c) pcm-ViT-magnitude, (d) from-scratch pcm-ViT-magnitude, (e) plain-ViT-inception, and (f) pcm-ViT-inception.}
    \label{fig:second_six_density_scattler}
\end{figure*}

\begin{figure*}[h]
    \centering
    \begin{minipage}[b]{0.8\columnwidth}
        \centering
        \includegraphics[width=\linewidth]{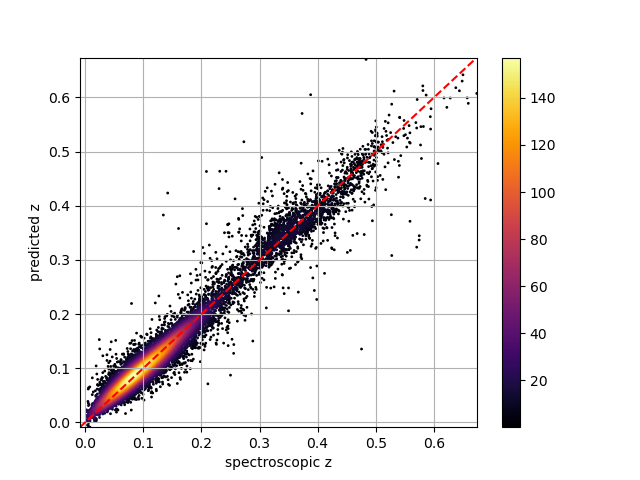}
        \caption*{(a)}
    \end{minipage}
    \begin{minipage}[b]{0.8\columnwidth}
        \centering
        \includegraphics[width=\linewidth]{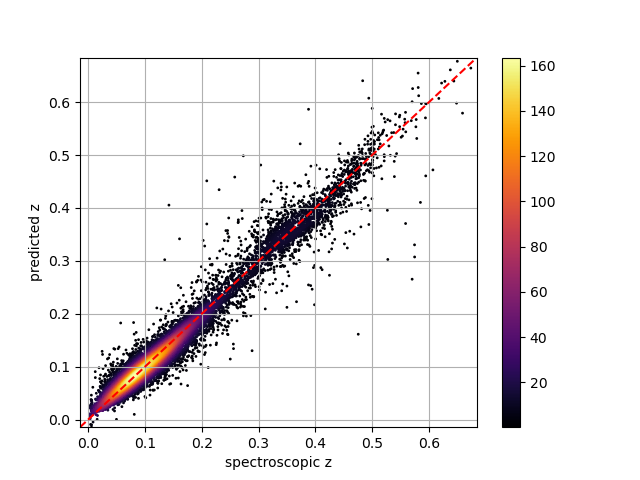}
        \caption*{(b)}
    \end{minipage}
    \caption{(a) Proposed plain-AstroMAE, and (b) Proposed pcm-AstroMAE.}
    \label{fig:proposed_AstroMAE_density_scattler}
\end{figure*}

In this section, we analyze the results and discuss the potential advantages and drawbacks of our approach. The performance metrics are summarized in Table \ref{tab:performance_metrics}. To further evaluate the performance of the predicted redshifts compared to their corresponding spectroscopic ground truths, we generated density scatter plots for all experiments. These plots are displayed in Fig. \ref{fig:first_six_density_scattler}, \ref{fig:second_six_density_scattler}, \ref{fig:proposed_AstroMAE_density_scattler}.

\textbf{Masked autoencoder provides valuable information for fine-tuning through unlabeled images:} As mentioned before, one reason behind pretraining is to extract general patterns from the data, which are not specifically associated with a single task. According to Table \ref{tab:performance_metrics}, the results of plain-ViT are much better compared to its from-scratch counterpart, as it has lower MSE, MAE, and Precision, and higher $R^2$. This clearly demonstrates the power of the pretrained encoder of AstroMAE in identifying valuable general patterns. 

\textbf{pcm-transformer can improve the lack of locality in plain-transformer:} pcm-ViT obtained better results in terms of most metrics compared to plain-ViT. However, the improvement is not as significant compared to the results obtained by from-scratch pcm-ViT versus from-scratch plain-ViT. This demonstrates that the PCM module can gather more local information related to redshift prediction in supervised learning compared to during pretraining.

\textbf{Inception-only redshift prediction is still more powerful than vision transformer models:} Results demonstrate that the Inception-only redshift prediction can extract more relevant features for redshift prediction. Based on research conducted by Si et al. \cite{ra1}, Inception modules can provide local information very well, including local edges and texture. This experiment shows that for redshift prediction, local information is more important than the global dependencies captured by transformer-based architectures, which include overall object structures.

\textbf{Vision transformer can increase the performance of the Inception-only redshift prediction:} Results of pcm-ViT-inception and plain-ViT-inception are remarkably better than the Inception-only redshift prediction. This demonstrates that for redshift prediction, in addition to the local information provided by the Inception modules, global dependency is necessary. Furthermore, the results show that pcm-ViT-inception achieves better outcomes compared to plain-ViT-inception.

\textbf{Magnitude block can improve results:} Results show that magnitude values corresponding to images can improve results significantly. These magnitudes are obtained by performing photometry on images. The improvement in results after adding the magnitude block demonstrates that both the Inception-only redshift prediction \cite{m8} and transformer-based models cannot gather these magnitudes from the images alone. For this reason, in our proposed fine-tuning architecture, the magnitude block is added to pcm-ViT-inception and plain-ViT-inception. The results demonstrate the capability of our fine-tuning architecture compared to other approaches.

\textbf{Proposed AstroMAEs Outperform Henghes et al. \cite{m8}: }
Both proposed AstroMAE models outperform Henghes et al. \cite{m8}. The key difference between the proposed AstroMAE and Henghes et al. \cite{m8} is the use of a transformer-based model in fine-tuning. This demonstrates that, in addition to local information, capturing global dependencies and general patterns in the data is crucial for accurate redshift prediction.

\begin{figure}[h]
  \centering
  \includegraphics[width=\columnwidth]{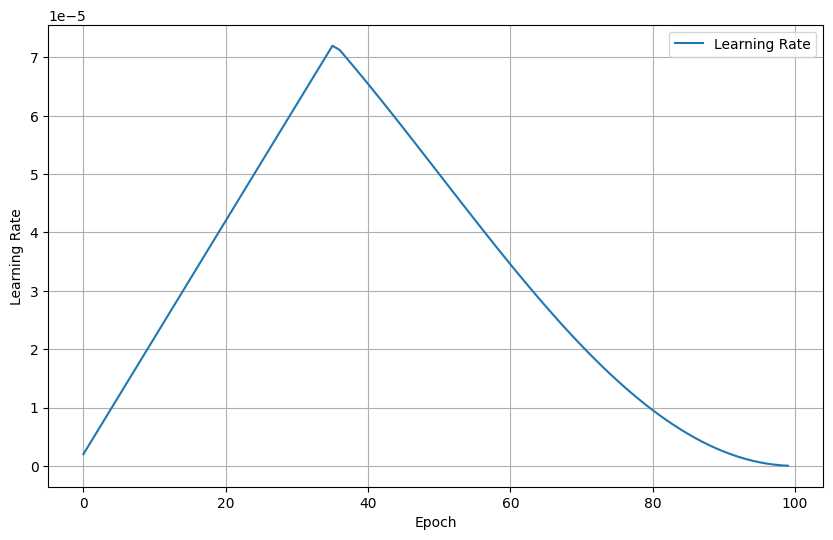}
  \caption{Learning rate for training during the second experiment.}
  \label{fig:finetuned_sec_lr}
\end{figure}

\subsection{Second Experiment}
In this experiment, we aim to evaluate the performance of pcm-AstroMAE and the baseline model \cite{m8}, utilizing 100\% of the data for both pretraining and fine-tuning. Although the baseline model is a supervised learning approach and could potentially benefit from the increased amount of labeled data, our results indicate that pcm-AstroMAE consistently outperforms the baseline. Table \ref{tab:performance_comparison} demonstrates the superiority of pcm-AstroMAE across evaluated metrics.

\begin{table}[h]
\centering
\caption{Comparison of Baseline Model and pcm-AstroMAE Performance using 100\% of data.}
% \textcolor{blue}{
\begin{tabular}{l c c}
\hline
\textbf{Metric} & \textbf{Baseline Model} & \textbf{pcm-AstroMAE} \\ \hline
MSE       & 0.00037 & \textbf{0.00033} \\ 
MAE       & 0.01302 & \textbf{0.01267} \\ 
Bias      & \textbf{-0.00157} & 0.00191 \\ 
Precision & 0.01192 & \textbf{0.01171} \\ 
R²        & 0.96899 & \textbf{0.97239} \\ \hline
\end{tabular}
\label{tab:performance_comparison}
\end{table}

\subsubsection{Training Configuration}
The configuration for pretraining remains consistent with the first experiment. For fine-tuning and training the baseline, the training scheduler adheres to the configuration illustrated in Figure \ref{fig:finetuned_sec_lr}. Additionally, Gaussian noise augmentation has been increased to 0.20 to further mitigate overfitting, leading to more stable training outcomes.

\section{Conclusions and Future Works}
In this paper, we employ a masked autoencoder—an efficient self-supervised learning method based on different transformer layers—for pretraining. To enhance the extraction of local information, we propose a novel hybrid fine-tuning method using inception modules for redshift prediction on the SDSS survey. Extensive experiments on various architectures constructed with vision transformers and CNNs demonstrate the lack of locality in transformer layers and the superiority of our method in addressing this issue. Based on the results, AstroMAE proves to be a successful redshift prediction method compared to the other methods tested in this paper.

In the next step, we aim to test our method on a broader range of downstream tasks to further demonstrate its capability and generality. We plan to extend the second experiment by evaluating all architectures mentioned in the first experiment using the full dataset. Additionally, we will conduct experiments to explore the effect of various mask ratios during pretraining on astrophysical image data. Furthermore, we intend to compare our methods with traditional approaches commonly used in astrophysics for redshift prediction.

\section{Acknowledgements}
We thank the University of Virginia Computer Science Department, the Biocomplexity Institute, and the Department of Energy Grant DE-SC0023452: "FAIR Surrogate Benchmarks Supporting AI and Simulation Research" for partial support.

\vspace{12pt}


\begin{thebibliography}{00}

\bibitem{int1}Hubble, E. (1929). A relation between distance and radial velocity among extra-galactic nebulae. Proceedings of the national academy of sciences, 15(3), 168-173.

\bibitem{int2}Dey, B., Andrews, B. H., Newman, J. A., Mao, Y. Y., Rau, M. M., \& Zhou, R. (2022). Photometric redshifts from SDSS images with an interpretable deep capsule network. Monthly Notices of the Royal Astronomical Society, 515(4), 5285-5305.

\bibitem{int3}Beck, R., Dobos, L., Budavári, T., Szalay, A. S., \& Csabai, I. (2016). Photometric redshifts for the SDSS Data Release 12. Monthly Notices of the Royal Astronomical Society, 460(2), 1371-1381.

\bibitem{int5}Fan, X., McGreer, I., Patej, A., Alvarez, A., Choi, Y., Jannuzi, B. T., ... \& Zaritsky, D. (2019). Overview of the DESI Legacy Imaging Surveys.

\bibitem{int6}Aihara, H., Arimoto, N., Armstrong, R., Arnouts, S., Bahcall, N. A., Bickerton, S., ... \& Yuma, S. (2018). The Hyper Suprime-Cam SSP survey: overview and survey design. Publications of the Astronomical Society of Japan, 70(SP1), S4.

\bibitem{int4}Salvato, M., Hasinger, G., Ilbert, O., Zamorani, G., Brusa, M., Scoville, N. Z., ... \& Zamojski, M. (2008). PHOTOMETRIC REDSHIFT AND CLASSIFICATION FOR THE XMM–COSMOS SOURCES. The Astrophysical Journal, 690(2), 1250.

\bibitem{int9}D’Isanto, A., \& Polsterer, K. L. (2018). Photometric redshift estimation via deep learning-generalized and pre-classification-less, image based, fully probabilistic redshifts. Astronomy \& Astrophysics, 609, A111.

\bibitem{adamw}Loshchilov, I., \& Hutter, F. (2017). Decoupled weight decay regularization. arXiv preprint arXiv:1711.05101.

\bibitem{m7}Pasquet, J., Bertin, E., Treyer, M., Arnouts, S., \& Fouchez, D. (2019). Photometric redshifts from SDSS images using a convolutional neural network. Astronomy \& Astrophysics, 621, A26.

\bibitem{int10}Rastegarnia, F., Mirtorabi, M. T., Moradi, R., Vafaei Sadr, A., \& Wang, Y. (2022). Deep learning in searching the spectroscopic redshift of quasars. Monthly Notices of the Royal Astronomical Society, 511(3), 4490-4499.
\bibitem{int11}Pasquet-Itam, J., \& Pasquet, J. (2018). Deep learning approach for classifying, detecting and predicting photometric redshifts of quasars in the Sloan Digital Sky Survey stripe 82. Astronomy \& Astrophysics, 611, A97.

\bibitem{int12}Sandeep, V. Y., Sen, S., \& Santosh, K. (2021, July). Analyzing and processing of astronomical images using deep learning techniques. In 2021 IEEE international conference on electronics, computing and communication technologies (CONECCT) (pp. 01-06). IEEE.

\bibitem{int13}Syarifudin, M. R. I., Hakim, M. I., \& Arifyanto, M. I. (2019, May). Applying deep neural networks (dnn) for measuring photometric redshifts from galaxy images: Preliminary study. In Journal of Physics: Conference Series (Vol. 1231, No. 1, p. 012013). IOP Publishing.

\bibitem{int14}Schuldt, S., Suyu, S. H., Cañameras, R., Taubenberger, S., Meinhardt, T., Leal-Taixé, L., \& Hsieh, B. C. (2021). Photometric redshift estimation with a convolutional neural network: NetZ. Astronomy \& Astrophysics, 651, A55.

\bibitem{intViT}Lin, J. Y. Y., Liao, S. M., Huang, H. J., Kuo, W. T., \& Ou, O. H. M. (2021). Galaxy morphological classification with efficient vision transformer. arXiv preprint arXiv:2110.01024.

\bibitem{int15}Jing, L., \& Tian, Y. (2020). Self-supervised visual feature learning with deep neural networks: A survey. IEEE transactions on pattern analysis and machine intelligence, 43(11), 4037-4058.
\bibitem{int16}Pathak, D., Krahenbuhl, P., Donahue, J., Darrell, T., \& Efros, A. A. (2016). Context encoders: Feature learning by inpainting. In Proceedings of the IEEE conference on computer vision and pattern recognition (pp. 2536-2544).

\bibitem{int17}Hayat, M. A., Stein, G., Harrington, P., Lukić, Z., \& Mustafa, M. (2021). Self-supervised representation learning for astronomical images. The Astrophysical Journal Letters, 911(2), L33.

\bibitem{int18}Lanusse, F., Parker, L., Golkar, S., Cranmer, M., Bietti, A., Eickenberg, M., ... \& Ho, S. (2023). AstroCLIP: Cross-Modal Pre-Training for Astronomical Foundation Models. arXiv preprint arXiv:2310.03024.

\bibitem{int19}Shen, G., Zou, Z., Luo, A. L., Hong, S., \& Kong, X. (2023). A Galaxy Morphology Classification Model Based on Momentum Contrastive Learning. Publications of the Astronomical Society of the Pacific, 135(1052), 104501.

\bibitem{int20}Stein, G., Harrington, P., Blaum, J., Medan, T., \& Lukic, Z. (2021). Self-supervised similarity search for large scientific datasets. arXiv preprint arXiv:2110.13151.

\bibitem{int21}Donoso-Oliva, C., Becker, I., Protopapas, P., Cabrera-Vives, G., Vishnu, M., \& Vardhan, H. (2023). ASTROMER-A transformer-based embedding for the representation of light curves. Astronomy \& Astrophysics, 670, A54.

\bibitem{int22}Schroff, F., Kalenichenko, D., \& Philbin, J. (2015). Facenet: A unified embedding for face recognition and clustering. In Proceedings of the IEEE conference on computer vision and pattern recognition (pp. 815-823).

\bibitem{int23}Chen, T., Kornblith, S., Norouzi, M., \& Hinton, G. (2020, November). A simple framework for contrastive learning of visual representations. In International conference on machine learning (pp. 1597-1607). PMLR.

\bibitem{int24}Chen, X., \& He, K. (2021). Exploring simple siamese representation learning. In Proceedings of the IEEE/CVF conference on computer vision and pattern recognition (pp. 15750-15758).


\bibitem{m4}He, K., Chen, X., Xie, S., Li, Y., Dollár, P., \& Girshick, R. (2022). Masked autoencoders are scalable vision learners. In Proceedings of the IEEE/CVF conference on computer vision and pattern recognition (pp. 16000-16009).


\bibitem{m1}Dosovitskiy, A., Beyer, L., Kolesnikov, A., Weissenborn, D., Zhai, X., Unterthiner, T., ... \& Houlsby, N. (2020). An image is worth 16x16 words: Transformers for image recognition at scale. arXiv preprint arXiv:2010.11929.

\bibitem{silu}Ramachandran, P., Zoph, B., \& Le, Q. V. (2017). Searching for activation functions. arXiv preprint arXiv:1710.05941.

\bibitem{gelu}Hendrycks, D., \& Gimpel, K. (2016). Gaussian error linear units (gelus). arXiv preprint arXiv:1606.08415.

\bibitem{m2}Xu, Y., Zhang, Q., Zhang, J., \& Tao, D. (2021). Vitae: Vision transformer advanced by exploring intrinsic inductive bias. Advances in neural information processing systems, 34, 28522-28535.

\bibitem{m3}Wang, D., Zhang, Q., Xu, Y., Zhang, J., Du, B., Tao, D., \& Zhang, L. (2022). Advancing plain vision transformer toward remote sensing foundation model. IEEE Transactions on Geoscience and Remote Sensing, 61, 1-15.

\bibitem{m5}Sun, X., Wang, P., Lu, W., Zhu, Z., Lu, X., He, Q., ... \& Fu, K. (2022). RingMo: A remote sensing foundation model with masked image modeling. IEEE Transactions on Geoscience and Remote Sensing.

\bibitem{m6}Szegedy, C., Liu, W., Jia, Y., Sermanet, P., Reed, S., Anguelov, D., ... \& Rabinovich, A. (2015). Going deeper with convolutions. In Proceedings of the IEEE conference on computer vision and pattern recognition (pp. 1-9).

\bibitem{m8}Henghes, B., Thiyagalingam, J., Pettitt, C., Hey, T., \& Lahav, O. (2022). Deep learning methods for obtaining photometric redshift estimations from images. Monthly Notices of the Royal Astronomical Society, 512(2), 1696-1709.

\bibitem{d1}Gunn, J. E., Carr, M., Rockosi, C., Sekiguchi, M., Berry, K., Elms, B., ... \& Brinkman, J. (1998). The sloan digital sky survey photometric camera. The Astronomical Journal, 116(6), 3040.

\bibitem{d2}Gunn, J. E., Siegmund, W. A., Mannery, E. J., Owen, R. E., Hull, C. L., Leger, R. F., ... \& Wang, S. I. (2006). The 2.5 m telescope of the sloan digital sky survey. The Astronomical Journal, 131(4), 2332.

\bibitem{d3}York, D. G., Adelman, J., Anderson Jr, J. E., Anderson, S. F., Annis, J., Bahcall, N. A., ... \& Yasuda, N. (2000). The sloan digital sky survey: Technical summary. The Astronomical Journal, 120(3), 1579.

\bibitem{d4}Ginsburg, A., Sipőcz, B. M., Brasseur, C. E., Cowperthwaite, P. S., Craig, M. W., Deil, C., ... \& Woillez, J. (2019). Astroquery: an astronomical web-querying package in Python. The Astronomical Journal, 157(3), 98.

% \bibitem{m8}Henghes, B., Thiyagalingam, J., Pettitt, C., Hey, T., \& Lahav, O. (2022). Deep learning methods for obtaining photometric redshift estimations from images. Monthly Notices of the Royal Astronomical Society, 512(2), 1696-1709.

\bibitem{TC1}He, T., Zhang, Z., Zhang, H., Zhang, Z., Xie, J., \& Li, M. (2019). Bag of tricks for image classification with convolutional neural networks. In Proceedings of the IEEE/CVF conference on computer vision and pattern recognition (pp. 558-567).
\bibitem{TC2}Xie, Z., Zhang, Z., Cao, Y., Lin, Y., Wei, Y., Dai, Q., \& Hu, H. (2023). On data scaling in masked image modeling. In Proceedings of the IEEE/CVF Conference on Computer Vision and Pattern Recognition (pp. 10365-10374).

\bibitem{TC3}Ba, J. L., Kiros, J. R., \& Hinton, G. E. (2016). Layer normalization. arXiv preprint arXiv:1607.06450.

\bibitem{mt1}Cohen, J. G., Hogg, D. W., Blandford, R., Cowie, L. L., Hu, E., Songaila, A., ... \& Richberg, K. (2000). Caltech faint galaxy redshift survey. x. a redshift survey in the region of the hubble deep field north. The Astrophysical Journal, 538(1), 29.
\bibitem{mt2}Ilbert, O., Arnouts, S., Mccracken, H. J., Bolzonella, M., Bertin, E., Le Fèvre, O., ... \& Vergani, D. (2006). Accurate photometric redshifts for the CFHT legacy survey calibrated using the VIMOS VLT deep survey. Astronomy \& Astrophysics, 457(3), 841-856.

\bibitem{ra1}Si, C., Yu, W., Zhou, P., Zhou, Y., Wang, X., \& Yan, S. (2022). Inception transformer. Advances in Neural Information Processing Systems, 35, 23495-23509.

\end{thebibliography}
\end{document}